\documentstyle[psfig,alife5]{article}

\newcommand{\smallgraph}[5]{\begin{figure*}\begin{center}\psfig{figure=#1.eps,width=#3in,height=#2in}\caption{#4}\label{f:#1}\end{center}\end{figure*}}

\begin{document}

\bibliographystyle{plain}

\title{Evolution of Neural Networks to Play the Game of Dots-and-Boxes\footnotemark[1]}

\author{Lex Weaver  and Terry Bossomaier\\Email: lex@cs.anu.edu.au, tbossomaier@csu.edu.au}
\maketitle

\begin{abstract}
{\em Dots-and-Boxes} is a child's game which remains analytically unsolved. We
implement and evolve artificial neural networks to play this game, evaluating them against simple heuristic players. Our networks do not evaluate or predict the final outcome of the game, but rather recommend moves at each stage. Superior generalisation of play by co-evolved populations is found, and a comparison made with networks trained by
back-propagation using simple heuristics as an oracle.
\end{abstract}

\renewcommand{\thefootnote}{\fnsymbol{footnote}}
\footnotetext[1]{L. Weaver and T. Bossomaier. Evolution of Neural Networks to Play the Game of Dots-and-Boxes. In {\em Alife V: Poster Presentations}, May 16-18 1996, pages 43-50.}
\renewcommand{\thefootnote}{\arabic{footnote}}

\section{Introduction}
The rapid growth in Artificial Life research brings us faster than we 
might ever have thought to the mystery of consciousness itself. There are 
still informed opinions from authors such as Searle~\cite{p:searle80} and Penrose~\cite{penrose89:enm,penrose94:sm} who believe that the 
workings of the human mind are fundamentally different to Turing 
computation. Yet the diversity and complexity of behaviour we see in ALife 
models  does little to reassure us that they are correct. Penrose has an 
illustrative argument~\cite[p46]{penrose94:sm} where he 
exhibits a chess problem, rapidly solved by novices yet incorrectly 
handled by Deep Thought~\cite{www:deepblue}, a chess computer of grandmaster rank.  The 
argument is that human analysis is somehow different. 

Not surprisingly, the best game playing programs build in all sorts 
of human-developed heuristics, which begs the question of what an 
artificial player can do by just evolving or learning. This question 
is the theme of the present paper. Tesauro~\cite{p:tesauro92} has demonstrated that a 
neural network can reach the top rank of backgammon players with little more than 
feedback as to which games it wins and which it loses. In the present
work we examine what capability a neural network can develop through
evolution alone, in playing a game which is superficially much simpler.
This game is sometimes referred to as \emph{Dots-and-Boxes}~\cite{p:holladay66,p:nowakowski91,www:dandbrules}. It is
described in detail in section~\ref{dotsandboxes}.

The game belongs to a class, some members of which have been completely solved,
others, of which this is one, have not \cite{p:meyniel&roudneff88}. Its interest to us is
that it embodies several levels of strategy and can be made arbitrarily 
small or large. The scalability of the game provides a platform on which
local strategies found valid in small but non-trivial examples of the game are also valid in multiple contexts on larger examples. It is envisaged that the propagation of such strategies to larger games may provide a means of quickly constructing a system capable of reasonable play on these large instances. Human players (usually
children: this is not a difficult game) have several different
strategies, discussed in section~\ref{dotsandboxes:strategies}. 
It is also advantageous that these  strategies can be ranked
according to the degree of \textit{temporal} lookahead they exhibit, and the extent to which position evaluation is restricted \textit{spatially}. 

Additionally,  we can not only clearly enunciate the strategies, but 
we can use simple heuristics to generate networks and 
training sets for supervised learning. Thus we can compare the 
performance of the genetic algorithm with clearly defined alternatives.

Our ALife player is a neural network. In the results discussed here,
the training of the network is accomplished through simple
evolution. In game strategies in general, feedback is a win/lose
alternative at the end of many moves. Direct feedback on a move by
move basis would require injecting move criteria, thus in some ways
training the net to behave in  the way we think that the game should
be played. However, in the present game  we can provide precise
feedback at every move against a player with a fixed lookahead (which, as discussed below, correlates well with human strategies).

\section{The Game of Dots and Boxes}
\label{dotsandboxes}

There are two forms of this game, one of which has been solved analytically~\cite{p:holladay66}. The other, which has one rule fewer,  is described below and has been analysed by Nowakowski~\cite{p:nowakowski91}. We concern ourselves only with this more general and unsolved form.

\subsection{Rules} 

The game is played between two players on a rectangular array of dots. A game is measured by the number of boxes the grid has horizontally and vertically, each is marked with a dot on every vertex.

\begin{figure}
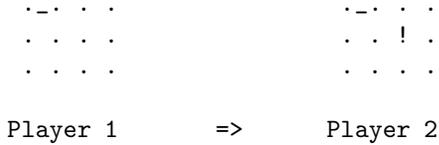

\begin{verbatim}
        ._. . .                ._. . .
        . . . .	               . . ! .
        . . . .	               . . . .
	
      	Player 1       =>      Player 2
\end{verbatim}
\caption{ The first two moves of a game.}
\label{f:game_start}
\end{figure}

The players take turns in making legal moves, where a legal move consists of joining two horizontally or vertically adjacent dots which were previously unjoined. A player may not ``pass'' on a move, that is, each move must involve the joining of two dots. As an example, two players might start the game with the moves shown in Figure \ref{f:game_start}.

When a player completes the fourth side of a box they score a point and {\em must} then make another move. {\em A player who can complete a box is not obliged to do so}. The analysis is simplified considerably if players are obliged to complete boxes, and this was the form solved by Holladay~\cite{p:holladay66}.

The game ends when there are no legal moves available, ie: there are no unjoined vertically or horizontally adjacent dots left. The winner is the player to have completed the most boxes (scored the most points). Since there may be an even number of boxes in the array of points, it is possible for the result to be a tie.

\subsection{Human Strategies} 
\label{dotsandboxes:strategies}
We can distinguish several levels of strategy in the game:

\begin{itemize}
	\item Level 0 --- Random Play. This corresponds to zero 
	\textit{temporal} levels of lookahead, and no \textit{spatial} analysis
	beyond whether a given move is legal.
	
	\item Level 1 --- Box Completion. This corresponds to one 
	\textit{temporal} level of 
	lookahead; the position of a hypothetical edge is tested to see if 
	it creates a box. This strategy has a \textit{spatial} 
	limitation wherein the 
	analysis is confined to a window of just one box at a time.
	Despite being simple, this strategy is quite successful, defeating a
	random opponent in 99.63\% of games (based on a sample of $200\:000$).
	
	\item Level 2 --- Third Side Avoidance. This 
	requires two lookahead steps, but is spatially still localised to one 
	box; moves which do not complete a box and would allow the opponent to
	are avoided. This strategy will defeat a random opponent in 
	99.69\% of games,
	and one using the box completion strategy in 83.83\% of games.
			
	\item Learning to Optimally Concede Boxes. The lookahead and locality 
	are now undefined. As the number of possible boxes increases, both 
	the level of lookahead and the spatial window need to increase in 
	order to see a set of multiple boxes. At the individual move level, 
	just taking one box at a time will collect all  
	available boxes. However, determining which set of multiple boxes 
	to allow the opponent to complete  requires greater 
	temporal and spatial analysis.
	
	\item  Learning to Ignore a Box.  To avoid giving 
	away a series of boxes, a box completion is deliberately spurned.
	This is illustrated in Figure~\ref{f:box_ignore}, where player 1's 
	only sure win is to move in the top-right corner and concede
	two boxes to player 2. This contrasts with the greedy approach
	of the previous strategies. Nowakowski~\cite{p:nowakowski91} refers to
	 this as the ``double-dealing endgame.'' Temporal and 
	spatial limitations for this strategy are again undefined.
	This strategy is often not even 
	learned by children playing the game, and to our knowledge is the 
	deepest strategy admitted.

\begin{figure}
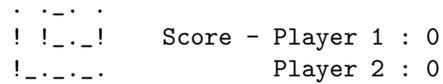


\begin{verbatim}
        . ._. .	
        ! !_._!    Score - Player 1 : 0
        !_._._.	           Player 2 : 0

\end{verbatim}
\caption{ Learning to ignore a box.}
\label{f:box_ignore}
\end{figure}
\end{itemize}

From both the evolution and neural network perspective, 
this last strategy is interesting, as it may involve 
{\em unlearning an existing strategy} to take a box whenever possible.

\section{Implementation as a Genetic Algorithm}
We have implemented a neural network which plays the game of Dots-and-Boxes on a board of size $3\times3$ (chosen to be non-trivial and avoid tied games). This size corresponds to the minimum spatial window which admits all the strategic components outlined in section \ref{dotsandboxes:strategies}. The weights of the network have been optimised using a genetic algorithm, with back-propagation being used as a comparative benchmark (see section~\ref{supervised-training}). The playing ability of the resulting networks was evaluated by play against heuristic players with zero or one level of look ahead (random and box completion strategies respectively).

Within the framework of the genetic algorithm, three main variants have been compared.

\paragraph{Direct Evolution} --- the population members are ranked according to how well they perform in a number of two game matches  against a heuristic opponent using a single strategy from those outlined in section \ref{dotsandboxes:strategies}, each player taking the role of player 1 once in a match.

\paragraph{Co-evolution : Round-Robin} --- the population members are ranked by their performance in a round-robin tournament of two game matches involving all phenotypes, each phenotype being matched against each other exactly once.

\paragraph{Co-evolution : Antibody/Antigen with Implicit Fitness Sharing} --- again two game matches are played, but match pairings are determined randomly and the rewards for defeating a given phenotype are shared amongst all phenotypes to achieve this. 
~\\
~\\

It is important to note that only in the first variant is the evolving 
population exposed to a heuristic player that could be used in evaluation. The other variants use the diversity of opponents within the population to bootstrap themselves out of ignorance. Implicit Fitness Sharing specifically attempts to maintain diversity within the population by reducing the reward for success against mediocre phenotypes~\cite{tr:darwenDTR}. Without this, it is possible for the population to cluster around attractors in the search space, with the fitness landscape defined by the population serving to keep it near the attractor.

\subsection{Game Management}
Game play is governed by a management module written in C++, which accepts, validates and processes moves, and maintains the ``official'' game record. Each player is required to have a C++ wrapper class which presents a common interface to the manager.

%\#\#\#Description of genetic code, fitness sharing operations (including 
%explanation of why fitness sharing is used) etc. Mechanisms of 
%playing the game etc.

\subsection{Choice of Network}
The network is a simple feed forward net, with one input node per 
edge (realised or potential) on the game board, one hidden unit per box and one output unit per edge. 
The networks are fully connected, allowing arbitrary spatial and 
temporal windows;  connections can of course have zero weight, but the 
present encoding discussed in section~\ref{encoding}, does not allow 
sparse networks to arise very easily. Each node uses a sigmoidal activation function, and has a threshold (negative or positive) specified as a weight on a
constant input of -1.

Whenever the network is required to make a move, it is presented with the encoded current game board on the input nodes. After processing, the output nodes are ranked in a descending order,
with the exclusion of those corresponding to illegal moves. The move 
corresponding to the highest ranked node is then passed back to the game
manager as the selected move. If several nodes have the same highest value, one is chosen at random. The encoding of the game board presented on the
input layer represents uncompleted edges as 0, and completed edges as 1.

For the game size we are working with, $3\times3$, the network has 24 input nodes, 9 hidden nodes, and 24 output nodes.

\subsection{Encoding of the Network} 
\label{encoding}
The networks are encoded in a genotype of 582 bytes (6 bits unused). A direct encoding is
used, which specifies every weight (and threshold) for each node in turn.
Each weight is encoded as a binary integer in 10 bits, the integer
representing one of $1\:024$ evenly spaced floating point values between 
-64.0 and +64.0 inclusive. All instances of this encoding generate legal networks so we have closure under the mutation and crossover operators, but
epistasis and aliasing problems are present. 

Since the activity of a node is dependent upon the weighted sum of its inputs,
 changing one of these weights through crossover or mutation will affect the relevant importance of the other weights. With hidden nodes being specified in
 250 bits, and output nodes in 100 bits, the potential for epistasis problems cannot be ignored.
 
Equivalent phenotypes can be encoded with the hidden units specified in any of 9!  orderings, yielding $362\:880$ equivalent encodings for each network. 

%\#\#\#Network encoding allows aliasing: different genotypes can produce 
%topologically identical phenotypes (after untwisting the net).

\subsection{Fitness Evaluation}
\label{fitness-evaluation}
Fitness evaluation when using direct evolution involves each phenotype network being played against the specified heuristic player in ten two-game matches. Networks are rewarded for winning games in each match, two games yielding one point, one game a half point, and no games no points\footnote{Every network was also given a small bonus fitness ($1\times10^{-6}$) to preclude a total population fitness of zero --- the proportional reproduction operator cannot cope with such unfit populations.}. The linear scaling modification described by Goldberg~\cite[pp.76-79]{b:goldberg89} is subsequently applied to the raw fitness, such that the best individual accounts for 10\% of population after proportional reproduction.

Fitness evaluation for co-evolution using the round-robin involves each phenotype playing each other once only, the winner of a two game match scoring one point, the loser no points, and a draw yielding half a point for each player.
Since co-evolution provides a ranking correct in order and separation for the current fitness landscape (which is determined by the present population), and we wished to observe the effect of this ranking, the scaling modification mentioned above was not applied.

Co-evolution using the antibody/antigen technique is non-deterministic. The phenotypes are generated, 25\% are randomly chosen as {\em antigens}, and for each of these, 33\% of the entire population are also randomly chosen as test {\em antibodies} who play the antigen in a two game match. Thirty-three points were made available to be shared proportionately between the antibodies with successes against a given antigen. An antibody which defeated the antigen in both games was awarded four shares, a win with a loss yielding only one share\footnote{Initial experiments indicated that this 4:1 ratio was significantly superior to the 2:1 to which we are accustomed.}$^{,}$\footnote{As with the traditional fitness function a small constant was always added to preclude a total population fitness of zero.}. Note that reward is given only to antibodies, not antigens. Antigens may  act as antibodies to other antigens and thus will receive a reward if successful. For the reasons given previously, fitness scaling was not used with this evaluation scheme.

\subsection{Generation of the New Population}  
\label{new-population}
All populations consisted of 100 individuals and were evolved using proportional reproduction, single-point crossover (probability 0.6), and bit-wise mutation (probability 0.005 or $0.000\:72$). Each operator supported single elitism\footnote{The highest ranked individual is guaranteed reproduction and remains unchanged by mutation and crossover.}.

\subsection{Heuristic Networks}
\label{s:heuristic}
A simple network has been derived which implements the level 1 strategy of section \ref{dotsandboxes:strategies}. It has a node for each box, which just looks 
at the four component edges. Suitable weights and threshold enable this 
node to signal when it has found a three-sided box and hence a 
good move. A four-sided box is also signalled but this would 
be an illegal move and is rejected at the move selection stage. 

Continuing in this vein,  two-sided and four-sided boxes 
could be rejected using a XOR-like network, involving an additional 
layer.  Both of these networks could be reached by evolution, but they 
are considerably sparser than the evolved architecture, and thus present a 
substantial challenge.

\subsection{Supervised Training}
\label{supervised-training}
We can train the network directly by a supervised learning algorithm 
such as back-propagation~\cite[pp.448-458]{b:winston92}.

Training sets were constructed by playing games between two players, one moving randomly, and the other using the level 1 box completion strategy implemented via a minimax search. Each time the level 1 player makes a move, the current game board and the move recommendations are recorded. Within the recommendations, moves are either flagged as illegal, or encoded with the number of boxes the player would complete if the move was made, usually 0 or 1. This scheme is consistent with the representation used by the networks in which desirable moves have higher output values. Approximately half of the raw training data so generated consists of game positions in which no move will complete a box. Training on these positions gives no benefit to the network, so they are filtered out before training commences. When calculating the error for a given training pair, moves flagged as illegal are assumed to have zero error, thus, the network only learns from legal moves, and the algorithm does not attempt to adjust weights to match outputs for an illegal move.

Assuming the effectiveness of the training 
algorithm, this method should produce a network whose performance 
approaches the level 1 strategy.

\section{Results} 
\label{results}
After some initial experiments to tune the parameters of the genetic algorithm, populations were evolved under each of the variants mentioned above. At regular intervals the entire population and the number of games played at that point were recorded. These population snapshots were later evaluated by playing at least 300 games between each population member and both level~0 and level~1 opponents. The success or failure of a variant could then be seen in terms of both the absolute proficiency of play attained against the benchmark opponents, and the speed with which this proficiency was attained. Proficiency of play is measured as the proportion of games won against the chosen opponent, whilst speed is measured in terms of the number of games played by the genetic algorithm to achieve a given level of proficiency.

\subsection{Level 0 Opponent}
\label{s:randomopponent}
Figure~\ref{f:random_opponent} illustrates the evolutionary progress made by the best individuals in each of four populations, each population being a typical example of its genetic algorithm variant. In this instance the mutation rate was 0.072\%, although the same relativities have been seen to hold for the higher rate of 0.5\%.

\smallgraph{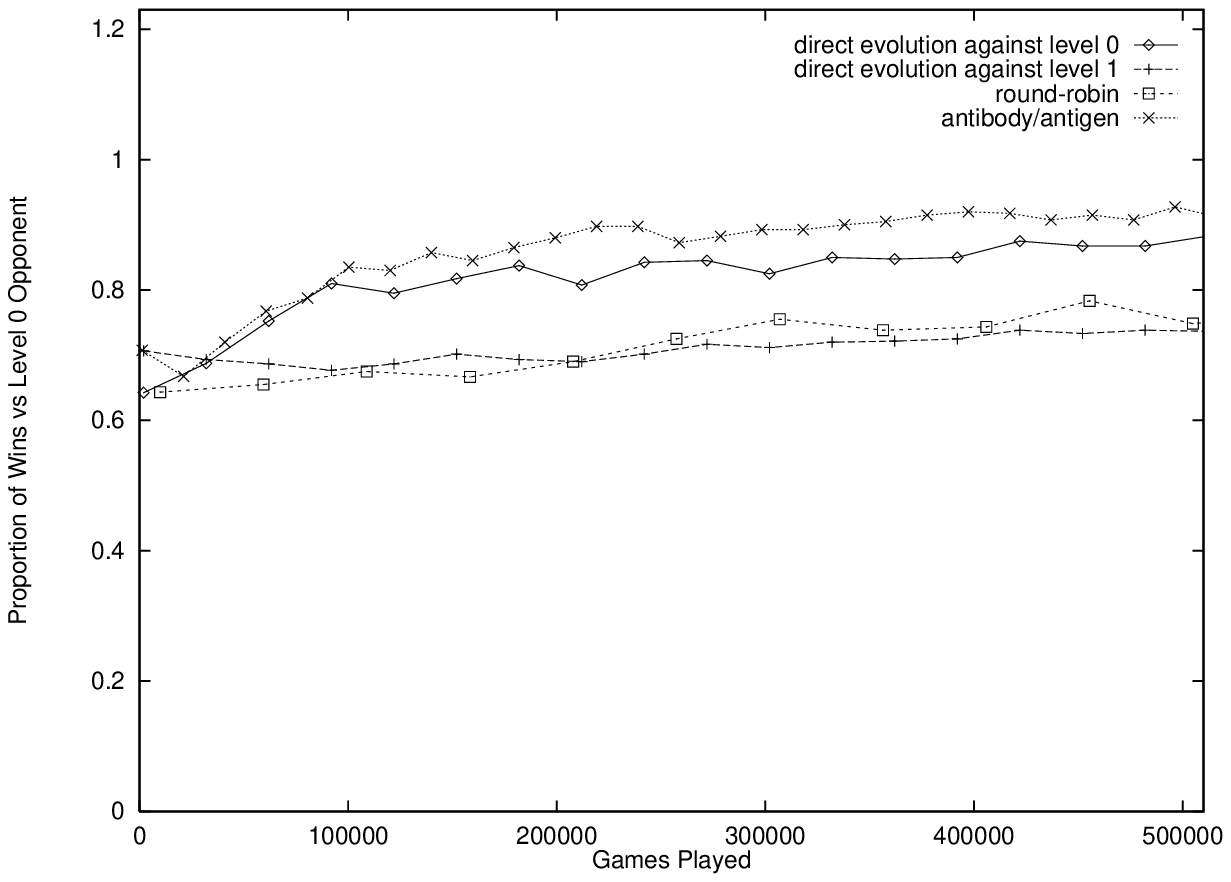}{3.8}{6}{Evolutionary success measured against a level 0 opponent}

The level 0 opponent is particularly easy to defeat. As indicated by the initial points for each population, the best of a random population is notably better than this opponent. However, of the four populations only two significantly increase their proficiency through evolution. The population evolved directly against the level 0 opponent, which specialises to defeat just this opponent, learnt quickly and plateaued at nearly nine wins in every ten games. However, co-evolution using the antibody/antigen technique, with implicit fitness sharing, yields populations which learn more quickly and plateau slightly higher, at just over nine wins in ten games. This contrasts with the other co-evolutionary variant which learns so slowly as to have improved by only one game in ten in Figure~\ref{f:random_opponent}. The population evolved directly against the level 1 opponent does not generalise well, showing even less improvement than the round-robin approach.

\subsection{Level 1 Opponent}
Figure~\ref{f:minimax_opponent} illustrates the evolutionary progress made by the best individuals in the same populations as discussed above, but measured against the level 1 heuristic opponent. This opponent is a far more proficient player than the level 0 opponent of the previous section, and the populations had far less success against it.

Again the population evolved directly against the evaluating opponent did well, this time being clearly the most successful. Of the others, only the population using the antibody/antigen technique managed an improvement, in the end being half as successful as the specialist population.

\smallgraph{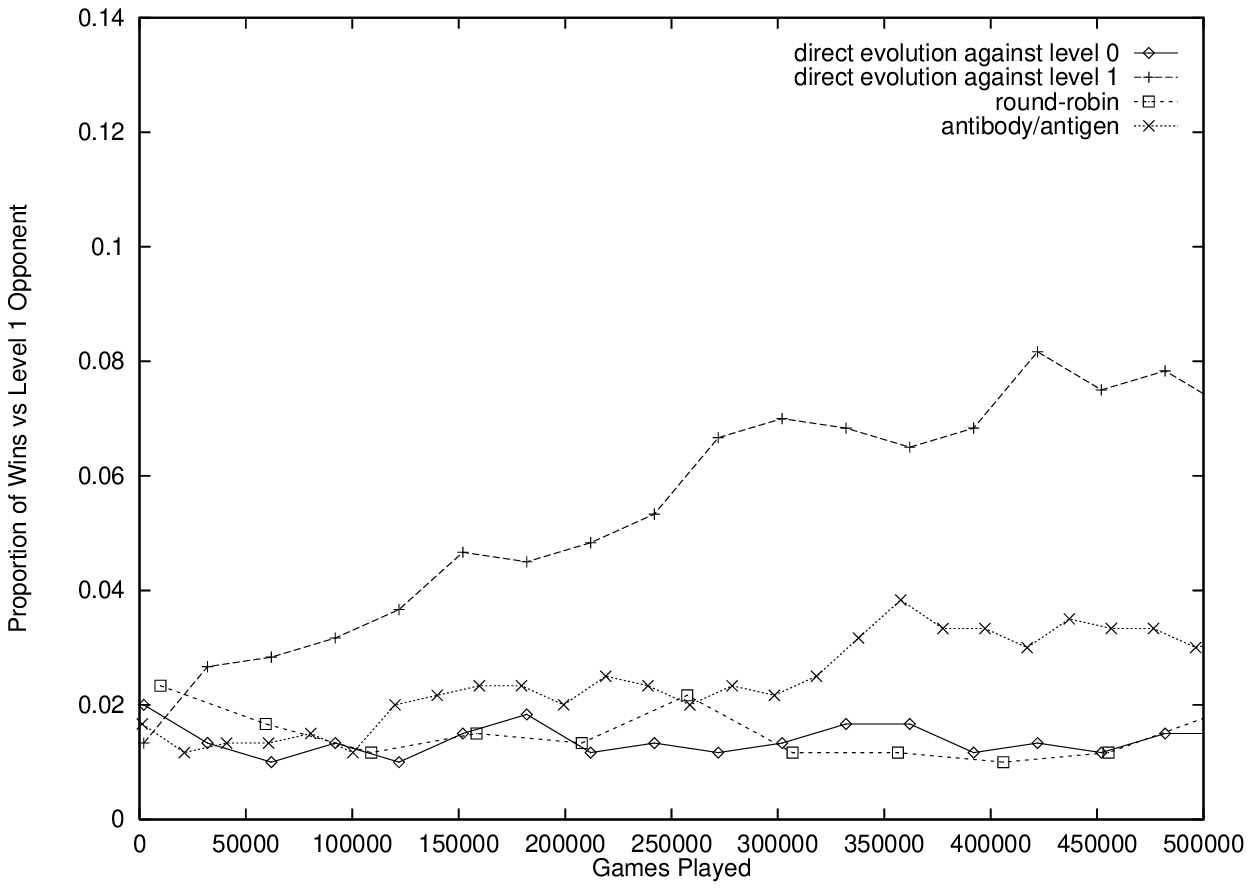}{3.8}{6}{Evolutionary success measured against a level 1 opponent}

\subsection{Performance of Supervised-Trained Nets} 
Using training sets generated as described in section~\ref{supervised-training}, attempts were made to train initially random networks, with the hope that their performance would approach that of a level 1 player. Figure \ref{f:backprop} shows a typical set of results, giving the mean and maximum performance of thirty runs spread across ten training sets ($800$ games yielding approximately $7\:250$ usable positions). In terms of CPU time, the cost of training these networks is about the same as evolving the populations discussed above, yet the best performed trained network is inferior to the best of each evolved populations. The training, in general, has a positive affect in the networks, but the rate of improvement is excruciatingly slow, and in the results shown may even have plateaued.

Evaluating the back propagation trained networks against the level 1 player yielded even less success. The mean performance was no better than random play, and the best networks still trailed the evolved populations.

Possible reasons for the bad performance of the back propagation trained networks are discussed in section~\ref{s:disc:supervised} below.

\smallgraph{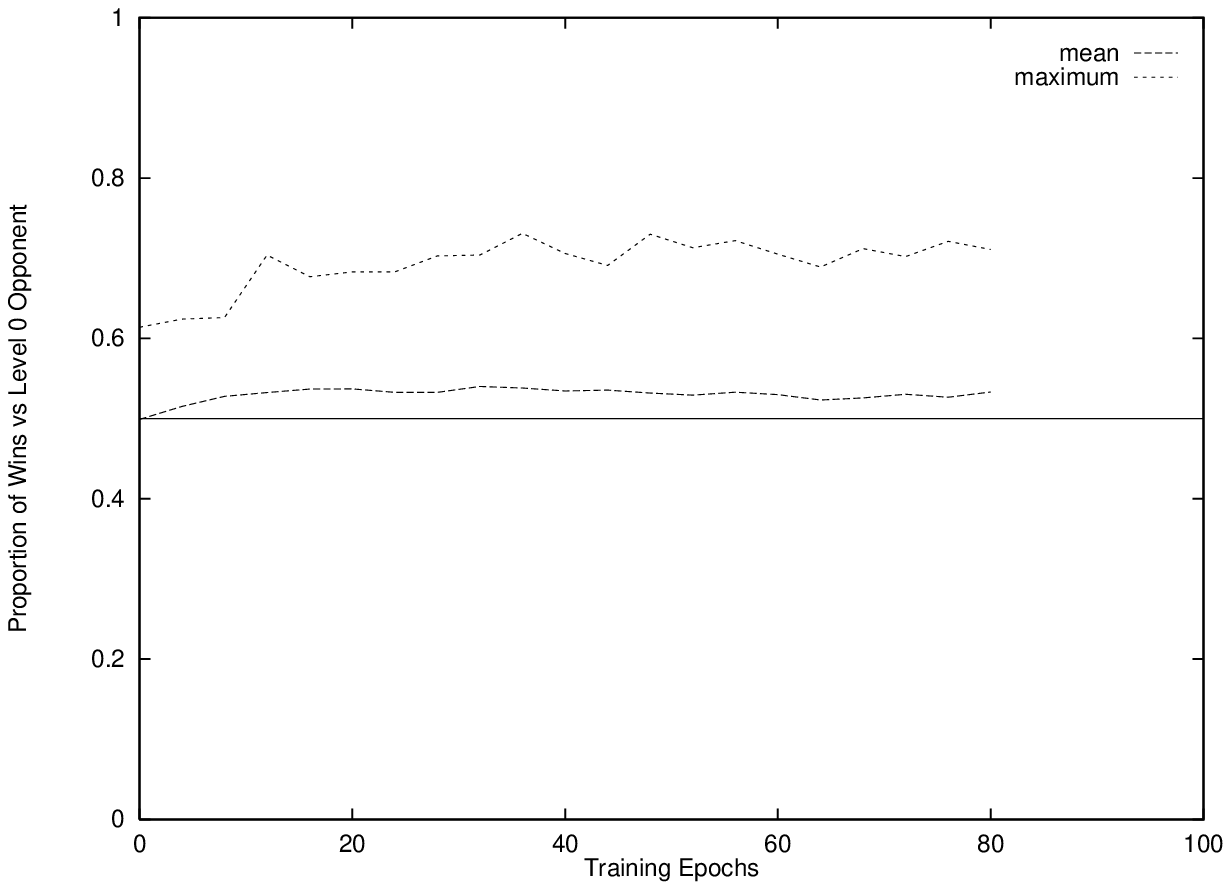}{3.8}{6}{Training success measured against a level 0 opponent}

\section{Discussion} 
\label{s:discussion}

The poor performances of both the evolved and trained networks against the level 1 opponent, contrasts with section~\ref{s:heuristic} in which we argue that a subset of the present architecture is capable of learning this strategy. The two learning techniques each have different problems pre-disposing them to poor performance.

\subsection{The Genetic Algorithm}
The genetic algorithm used only evolves the weights and thresholds of the neural network, not the architecture. The network outlined in section~\ref{s:heuristic} is quite sparse, only requiring each hidden node to have four non-zero weights and each output node one or two. With 432 weights encoded in the genotype, and the zero value being but one bit pattern of $1\:024$ possible patterns for each weight, it can be seen that a network as sparse as the one described is unlikely to form.

The population evolved through co-evolution using the antibody/antigen technique with implicit fitness sharing performed well against both the level 0 and level 1 opponents, indicating good generalisation in its playing proficiency. This property was not exhibited by any of the other populations evolved, and validates the use of implicit fitness sharing as a means of encouraging strategy diversification within evolving populations.

The use of the round-robin tournament in co-evolution suffers from two problems. As mentioned earlier, there is no penalty for individuals clustering around a single attractor in the search space. The technique also requires a large number of games to be played for each generation. In the populations evolved, each of 100 individuals, the round-robin required $9\:900$ games per generation, whilst the implicit fitness sharing needed only $1\:650$, and the traditional fitness function $2\:000$. Thus, the round-robin approach explores the search space six times more slowly than the implicit fitness sharing technique, and does so based upon fitness rankings which do not encourage the strategic diversity needed to gain generalised proficiency.

\subsection{Supervised Learning}
\label{s:disc:supervised}
The networks which underwent supervised training may have suffered for several reasons. The training may have been inadequate, or there may be inherent problems in the back-propagation algorithm making it unsuitable to this class of problem.

In the context of all possible game positions on a $3\times 3$ grid, the $7\:000$ or so positions trained upon appears minute. Even with the knowledge that less than half of the positions encountered in a game include the potential for a box to be completed, this number is small. However, experiments run using both fewer and more training positions did not yield significantly different results. Similarly, experiments using different learning rates did not produce notably different results.

The more likely cause for the poor performance of networks under supervised training is an interaction between their high connectivity, the back-propagation algorithm, and the particular application at hand. If a network was perfectly trained by some method, it would respond to a snapshot of the game state by ensuring that all of the equally beneficial best moves would receive the highest score on their output nodes. What this score is, or what the other lower scores are does not matter. What is important in the network output, is the relativities between the output nodes, {\em not} their actual values. Back-propagation works by propagating the difference between actual and expected outputs back through the connections, incrementally altering weights to reduce the error. Thus, back-propagation will achieve the desired relativities between the output nodes, only by working towards achieving the expected output values. In short, it can only give us a solution by solving a problem far harder than what is needed. Additionally, in the short term, tackling the harder problem may not improve performance on the easier.

This harder problem is actually quite difficult, given the highly connected architecture we are dealing with, and the discrete nature of our training data. Typically, an output pattern in the training data will have only one node with a positive value, the others being zero or flagged as illegal moves. If several of the zero valued nodes have errors (to be expected with a random initial network), then weight adjustments to correct these will dominate any adjustments related to the lone positive node. This is particularly relevant in the context of our fully connected network, where all hidden and input nodes contribute to each erroneous output. Thus, learning may be excruciatingly slow, as the network first tries to adjust 465 weights and thresholds to reduce errors associated with the zero values in each output pattern and then considers the lone positive value. 

We feel that supervised training is worthwhile only as a benchmark, since for complex games we do not, in general, know the best move in any situation. However, it is remarkable that the genetic algorithm should succeed where a more
direct hill-climbing approach should fail.

\subsection{Future Work}
The evolution of network architecture may offer a solution to the problems associated with weight evolution on optimally sparse networks. We plan to investigate this and combine it with training of the weights, leading to a comparison of Lamarckian Evolution and the Baldwin Effect~\cite{p:baldwin1896} on this application. 
 
A significant problem yet to be tackled is the propagation of strategies learnt on smaller instances of a game to larger instances. Traditionally defined network architectures which are a fixed and immutable whole, are not amenable to spatial and temporal replication of discovered strategies. Indeed with these architectures, it is usually necessary for strategies to be discovered several times to be fully applied across the network. In the current application, the box completion strategy would need to be discovered once for each box. Modular network construction~\cite{p:bossomaier&snoad95}, with re-usable modules implementing simple strategies being developed and propagated throughout an existing network and/or to larger instances, is a possible solution.

\section{Conclusions}

\emph{Dots-and-Boxes} is a simple game which is well suited to the research at hand. It has several useful properties, including clearly defined strategies and scalability. The strategies are characterised by differing temporal and spatial windows, giving rise to easily classified heuristic players. Together with random players these are useful benchmarks for evaluating the proficiency of our networks. The ability to construct, by inspection, networks which exhibit certain strategies also gives us a firm lower bound on the capabilities of different network architectures.

Our experiments indicate that genetic algorithms are a valid means of training game-playing neural networks, and can be superior to supervised learning techniques. Specifically, we have demonstrated that co-evolution using an antibody/antigen technique with implicit fitness sharing is significantly more efficient than the traditional round-robin tournament, and produces populations whose game-playing proficiency generalises well to new opponents. More effective feed-forward architectures may exist for back-propagation training. Our concern here has been to compare training via a conventional algorithm with the genetic algorithm. The latter appears far more robust.

Section~\ref{s:heuristic} indicates that our architecture is capable of at least the level 1 box completion strategy. This strategy requires a sparse network which will not easily arise in our present system. The failure of both supervised and unsupervised learning techniques to achieve this indicates a need for dynamic architectures.  We plan to investigate the evolution of network architecture as a solution to this problem. We also hope to implement a temporal difference learning system for the present application, both as a comparison and possibly a component of a larger neuro-genetic learning system.

\bibliography{dandb1}

\end{document}